\documentclass[runningheads]{llncs}

\usepackage{times,helvet,courier}
\usepackage[utf8]{inputenc}
\usepackage{amsmath}
\usepackage{amssymb}
\usepackage{microtype}
\usepackage{url}
\newcommand{\comment}[1]{} % \usepackage{comments}
\newcommand{\citeauthor}[1]{\cite{#1}}
\newcommand{\citeyear}[1]{\cite{#1}}

\usepackage{multicol}

\usepackage{listings}
\input{fitch.sty}

\newcommand\bcmdtab{\noindent\bgroup\tabcolsep=0pt%
  \begin{tabular}{@{}p{10pc}@{}p{20pc}@{}}}
\newcommand\ecmdtab{\end{tabular}\egroup}

\newcommand{\At}{\ensuremath{\mathcal{A}}} % {\ensuremath{\mathit{At}}}
\newcommand{\eqdef}{%
  \mathrel{\vbox{\offinterlineskip\ialign{%
    \hfil##\hfil\cr%
    $\scriptscriptstyle\mathrm{def}$\cr%
    \noalign{\kern1pt}%
    $=$\cr%
    \noalign{\kern-0.1pt}%
}}}}

\newcommand\tuple[1]{\langle #1 \rangle}

\def\N5{{\cal N}_5}
\def\X5{{\cal X}_5}

\def\qed{\hfill$\Box$}
\newtheorem{observation}{Observation}

\newcommand{\SM}[1]{\ensuremath{\mathit{SM}[#1]}}
\newcommand{\SMR}[2]{\ensuremath{\mathit{SM}_{#2}[#1]}}

\newcommand{\auxatom}[1]{\ensuremath{x_{#1}}}
\newcommand{\negocc}[1]{\ensuremath{\mathit{N}({#1})}}
\newcommand{\translation}[1]{\ensuremath{#1^{\star}}}
\newcommand{\auxatm}[1]{\ensuremath{\bar{#1}}}

\newcommand{\sysfont}{\textit}

\newcommand{\anthem}{\sysfont{anthem}}

\newcommand{\clingo}{\sysfont{clingo}}

\newcommand{\dlv}{\sysfont{dlv}}

\allowdisplaybreaks{}

\begin{document}

\title{Answer Set Programming Made Easy}

\author{%
  Jorge Fandinno \and
  Seemran Mishra \and
  Javier Romero \and
  Torsten Schaub}

\institute{University of Potsdam, Germany}

\author{Jorge Fandinno\inst{1,2} \and
Seemran Mishra\inst{2} \and
Javier Romero\inst{2} \and
Torsten Schaub\inst{2}}

\institute{University of Nebraska at Omaha, USA \and
University of Potsdam, Germany}

\maketitle

\begin{abstract}
  We take up an idea from the folklore of Answer Set Programming (ASP),
  namely that choices, integrity constraints along with a restricted rule format
  is sufficient for ASP.
  We elaborate upon the foundations of this idea in the context of the logic of Here-and-There and show how it can be derived
  from the logical principle of extension by definition.
  We then provide an austere form of logic programs that may serve as a normalform for logic programs similar to conjunctive normalform in
  classical logic.
  Finally, we take the key ideas and propose a modeling methodology for ASP beginners and illustrate how it can be used.
\end{abstract}

%%% Local Variables:
%%% mode: latex
%%% TeX-master: "paper"
%%% End:

\section{Introduction}\label{sec:introduction}

Many people like Answer Set Programming (ASP~\cite{lifschitz02a})
because its declarative approach frees them from expressing any procedural information.
In ASP,
neither the order of rules nor
the order of conditions in rule antecedents or consequents matter
and thus leave the meaning of the overall program unaffected.
Although this freedom is usually highly appreciated by ASP experts,
sometimes laypersons seem to get lost without any structural guidance
when modeling in ASP.

We address this issue in this (preliminary) paper and
develop a methodology for ASP modeling that targets laypersons,
such as biologists, economists, engineers, and alike.
As a starting point, we explore an idea put forward by Ilkka Niemel{\"{a}} in~\cite{niemela08b}, although already present in~\cite{felomisu93a,grsaza01a} as well as the neighboring area of Abductive Logic Programming~\cite{eshkow89a,denkak02a}.
To illustrate it, consider the logic program encoding a Hamiltonian circuit problem in Listing~\ref{lp:ham}.
%
% --------------------------------------------------------------------------------
\begin{lstlisting}[float=ht,caption={A logic program for a Hamiltonian circuit problem},label=lp:ham]
node(1..4).           start(1).
edge(1,2). edge(2,3). edge(2,4). edge(3,1).
edge(3,4). edge(4,1). edge(4,3).

{ hc(V,U) } :- edge(V,U).
reached(V) :- hc(S,V), start(S).
reached(V) :- reached(U), hc(U,V).
:- node(V), not reached(V).
:- hc(V,U), hc(V,W), U!=W.
:- hc(U,V), hc(W,V), U!=W.
\end{lstlisting}
% --------------------------------------------------------------------------------
%
Following good practice in ASP,
the problem is separated into the specification of the problem instance in lines~1-3
and the problem class in lines~5-10.
This strict separation, together with the use of facts for problem instances,
allows us to produce
uniform\footnote{A problem encoding is \textit{uniform}, if it can be used to solve all its problem instances.}
and
elaboration tolerant\footnote{A formalism is \emph{elaboration tolerant} if it is convenient
  to modify a set of facts expressed in the formalism to take into account new phenomena or changed circumstances~\cite{mccarthy98a}.}
specifications.
Building upon the facts of the problem instance,
the actual encoding follows the guess-define-check methodology of ASP.
A solution candidate is guessed in Line~5,
analyzed by auxiliary definitions in Line~6 and~7,
and finally checked through integrity constraints in lines~8-10.

A closer look reveals even more structure in this example.
From a global perspective,
we observe that the program is partitioned into
facts, choices, rules, and integrity constraints,
and in this order.
From a local perspective,
we note moreover that the predicates in all rule antecedents are defined beforehand.
This structure is not arbitrary and simply follows the common practice
that concept formation is done linearly by building concepts on top of each other.
Moreover, it conveys an intuition on how a solution is formed.
Importantly, such an arrangement of rules is purely methodological and has no impact on the meaning
(nor the performance\footnote{Shuffling rules in logic programs has an effect on performance since it affects tie-breaking during search;
  this is however unrelated to the ordering at hand.}) of the overall program.
From a logical perspective,
it is interesting to observe that the encoding refrains from using negation explicitly,
except for the integrity constraints.
Rather this is hidden in Line~5, where the choice on \lstinline{hc(V,U)} amounts
to the disjunction \lstinline[mathescape]{hc(V,U)$\,\vee\,\neg$hc(V,U)},
an instance of the law of the excluded middle.
Alternatively, \lstinline{hc(V,U)} can also be regarded as an abducible that may or may not be added to a program,
as common in Abductive Logic Programming.

Presumably motivated by similar observations,
Ilkka Niemelä already argued in~\citeyear{niemela08b} in favor of an ASP base language based on choices, integrity constraints, and stratified negation.
\footnote{This concept eliminates the (problematic) case of recursion through negation.}
We also have been using such an approach
when initiating students to ASP as well as teaching laypersons.
Our experience has so far been quite positive and we believe that a simple and more structured approach helps to
get acquainted with posing constraints in a declarative setting.

We elaborate upon this idea in complementary ways.
First of all, we lift it to a logical level to investigate its foundations and identify its scope.
Second, we want to draw on this to determine a syntactically restricted\comment{irreducible} subclass of logic programs
that still warrants the full expressiveness of traditional ASP.
Such a subclass can be regarded as a normalform for logic programs in ASP.
This is also interesting from a research perspective since it allows scientists to initially develop their theories in a restricted setting without
regarding all corner-cases emerging in a full-featured setting.
And last but not least, inspired by this,
we want to put forward a simple and more structured modeling methodology for ASP that aims at beginners and laypersons.

%%% Local Variables:
%%% mode: latex
%%% TeX-master: "paper"
%%% End:

\section{Background}\label{sec:background}

The logical foundations of ASP rest upon the logic of Here-and-There (HT~\cite{heyting30a})
along with its non-monotonic extension, Equilibrium Logic~\cite{pearce06a}.

We start by defining the monotonic logic of \emph{Here-and-There} (HT).
Let $\At$ be a set of atoms.
A \emph{formula} $\varphi$ over \At\ is an expression built with the grammar:
\[
\varphi ::= a \mid \bot \mid \varphi \wedge \varphi \mid \varphi \vee \varphi \mid \varphi \to \varphi
\]
for any atom $a\in \At$.
We also use the abbreviations:
\(
\neg \varphi \eqdef (\varphi \to \bot)
\),
\(
\top \eqdef \neg \bot
\),
and
\(
\varphi \leftrightarrow \psi \eqdef (\varphi \to \psi) \wedge (\psi \to \varphi)
\).
Given formulas $\varphi, \alpha$ and $\beta$,
we write $\varphi[\alpha/\beta]$ to denote the uniform substitution of all occurrences of formula $\alpha$ in $\varphi$ by~$\beta$.
This generalizes to the replacement of multiple formulas in the obvious way.
As usual, a \emph{theory} over \At\ is a set of formulas over \At.
We sometimes understand finite theories as the conjunction of their formulas.

An \emph{interpretation} over \At\ is a pair $\tuple{H,T}$ of atoms (standing for ``here'' and ``there'', respectively)
satisfying $H \subseteq T \subseteq \At$.
An interpretation is \emph{total} whenever $H=T$.
An interpretation $\tuple{H,T}$ \emph{satisfies} a formula $\varphi$, written $\tuple{H,T} \models \varphi$,
if the following conditions hold:
\[
  \begin{array}{rcll}
    \tuple{H,T} & \models & p                     & \text{if } p \in H\\
    \tuple{H,T} & \models & \varphi \wedge \psi   & \text{if } \tuple{H,T} \models \varphi \text{ and } \tuple{H,T}  \models  \psi\\
    \tuple{H,T} & \models & \varphi \vee \psi     & \text{if } \tuple{H,T} \models \varphi \text{ or } \tuple{H,T}  \models  \psi\\
    \tuple{H,T} & \models & \varphi\! \to \! \psi & \text{if } \tuple{H',T} \not\models \varphi \text{ or } \tuple{H',T} \models \psi \text{ for both } H'\in\{H,T\}
  \end{array}
\]
A formula $\varphi$ is \emph{valid}, written $\models \varphi$,
if it is satisfied by all interpretations.
An interpretation $\tuple{H,T}$ is a \emph{model} of a theory $\Gamma$, written $\tuple{H,T} \models \Gamma$, if $\tuple{H,T}\models \varphi$ for all $\varphi \in \Gamma$.

Classical entailment is obtained via the restriction to total models.
Hence, we define the classical satisfaction of a formula $\varphi$ by an interpretation $T$,
written $T\models\varphi$, as $\tuple{T,T}\models\varphi$.

A total interpretation $\tuple{T,T}$ is an \emph{equilibrium model} of a theory $\Gamma$ if $\tuple{T,T}$ is a model of $\Gamma$ and there is no other model $\tuple{H,T}$ of $\Gamma$ with $H \subset T$.
In that case, we also say that $T$ is a \emph{stable model} of~$\Gamma$.
We denote the set of all stable models of~$\Gamma$ by $\SM{\Gamma}$
and use $\SMR{\Gamma}{V} \, \eqdef \, \{\, T \cap V \mid T \in \SM{\Gamma}\,\}$
for their projection onto some vocabulary $V \subseteq \At$.

Since ASP is a non-monotonic formalism,
it may happen that two different formulas share the same equilibrium models
but behave differently in different contexts.
The concept of \emph{strong equivalence} captures the idea that two such formulas have the same models regardless of any context.
More precisely,
given two theories $\Gamma$ and $\Pi$ and a set $V \subseteq \At$ of atoms,
we say that $\Gamma$ and $\Pi$ are \emph{\mbox{$V$-strongly} equivalent}~\cite{agcafapepevi19a},
written $\Gamma \cong_V \Pi$,
if
$\SMR{\Gamma \cup \Delta}{V} = \SMR{\Pi \cup \Delta}{V}$
for any theory $\Delta$ over $\At'$ such that $\At' \subseteq V$.
For formulas~$\varphi$ and~$\psi$,
we write~$\varphi \cong_V \psi$
if $\{ \varphi \} \cong_V \{ \psi \}$.

A \emph{rule} is a (reversed) implication of the form
\begin{align}\label{rule}
l_1\vee\dots\vee l_m\leftarrow l_{m+1}\wedge\dots\wedge l_n
\end{align}
where each $l_i$ is a literal, that is, either an atom or a negated atom, for $1\leq i\leq n$.
If $n=1$, we refer to the rule as a \emph{fact} and write it as $l_1$ by dropping the trailing implication symbol.
A rule is said to be \emph{normal} whenever $m=1$ and $l_1$ is an atom.
A negation-free normal rule is called \emph{definite}.
An \emph{integrity constraint} is a rule with $m=0$ and equivalent to
\(
\bot\leftarrow l_{m+1}\wedge\dots\wedge l_n
\).
Finally, the law of the excluded middle $a\vee\neg a$ is often represented as $\{a\}$ and called a \emph{choice}.
Accordingly, a rule with a choice on the left-hand side is called a \emph{choice rule}.
A \emph{logic program} is a set of rules.
It is called \emph{normal}, if it consists only of normal rules and integrity constraints,
and \emph{definite} if all its rules are definite.

%%% Local Variables:
%%% mode: latex
%%% TeX-master: "paper"
%%% End:

\section{Logical Foundations}

We begin by investigating the logical underpinnings of the simple format of logic programs discussed in the introductory section.
Although the discussion of the exemplary logic program has revealed several characteristic properties,
not all of them can be captured in a logical setting, such as order related features.
What remains is the division of the encoding into facts, rules, choices, and integrity constraints.
In logical terms,
the first two amount to negation-free formulas,
choices are instances of the law of the excluded middle,
and finally integrity constraints correspond to double-negated formulas in HT.
While the first two types of formulas are arguably simpler because of their restricted syntax,
the latter's simplicity has a semantic nature and is due to the fact that in HT double negated formulas can be treated as in classical logic.

In what follows,
we show that any formula can be divided into a conjunction of corresponding subformulas.
This conjunction is strongly equivalent (modulo the original vocabulary) to the original formula and the translation can thus also be applied to substitute subformulas.
Interestingly, the resulting conjunction amounts to a conservative extension of the original formula and
the underlying translation can be traced back to the logical principle of extension by definition,
as we show below.

To this end,
we associate with each formula~$\varphi$ over \At\ a new propositional atom~\auxatom{\varphi}.
We then consider defining axioms of the form
\(
(\auxatom{\varphi} \leftrightarrow \varphi)
\).
We can now show that replacing any subformula $\varphi$ by \auxatom{\varphi} while adding a corresponding defining axiom amounts to a conservative
extension of $\psi$.
%
% --------------------------------------------------------------------------------
\begin{proposition}\label{prop:extension:definition}
  Let $\psi$ and $\varphi$ be formulas over \At\ and $\auxatom{\varphi} \not \in \At$.

  Then,
  \(
  \psi \cong_{\At} (\psi[\varphi/\auxatom{\varphi}] \wedge (\varphi \leftrightarrow \auxatom{\varphi}))
  \).
\end{proposition}
% --------------------------------------------------------------------------------
%
 \comment{T: comment on short proof and relation to Vladimir}
Moreover, we get a one-to-one correspondence between the stable models of both formulas.
%
% --------------------------------------------------------------------------------
\begin{proposition}\label{prop:extension:definition:one:one}
  Let $\psi$ and $\varphi$ be formulas over \At\ and $\auxatom{\varphi} \not \in \At$.

  \begin{enumerate}
  \item If $T\subseteq\At$ is a stable model of $\psi$,
    then $T \cup \{\auxatom{\varphi} \mid T \models \varphi \}$ is a stable model of
    $(\psi[\varphi/\auxatom{\varphi}] \wedge (\varphi \leftrightarrow \auxatom{\varphi}))$.
  \item If $T\subseteq (\At \cup \{\auxatom{\varphi}\})$ is a stable model of
    $(\psi[\varphi/\auxatom{\varphi}] \wedge (\varphi \leftrightarrow \auxatom{\varphi}))$,
    then $T\cap\At$ is a stable model of $\psi$.
  \end{enumerate}
\end{proposition}
% --------------------------------------------------------------------------------
%
Clearly, the above results generalize from replacing and defining a single subformula $\varphi$ to several such subformulas.

With this, we can now turn our attention to negated subformulas:
Given a formula~$\psi$, let $\negocc{\psi}$ stand for the set of all maximal negated subformulas occurring in $\psi$.
This leads us to the following variant of Proposition~\ref{prop:extension:definition}.
%
% --------------------------------------------------------------------------------
\begin{corollary}\label{cor:extension:definition:negated}
  Let $\psi$ be a formula over \At\ and $\auxatom{\varphi} \not \in \At$.

  Then,
  \(
  \psi
  \cong_{\At}
  \psi\big[\varphi/\auxatom{\varphi}\mid\varphi\in\negocc{\psi}\big]
  \wedge
  \bigwedge_{\varphi\in\negocc{\psi}}(\varphi\leftrightarrow\auxatom{\varphi})
  \).
\end{corollary}
% --------------------------------------------------------------------------------

Given that we exclusively substitute negated subformulas,
we can actually treat the defining axiom as in classical logic.
This is because in HT, we have
\(
\tuple{H,T} \models\neg \varphi
\)
iff
(classically)
\(
T \models\neg \varphi
\).
The classical treatment of the defining axiom is then accomplished by replacing
\(
(\varphi\leftrightarrow\auxatom{\varphi})
\)
by
\(
\neg \neg (\varphi\leftrightarrow\auxatom{\varphi})
\)
and
\(
(\neg\auxatom{\varphi}\vee\auxatom{\varphi})
\).
This results in the following decomposition recipe for formulas.
%
% --------------------------------------------------------------------------------
\begin{definition}\label{def:translation}
  Let $\psi$ be a formula over \At\ and $\auxatom{\varphi} \not \in \At$.

  Then, we define
  \[%\textstyle
    \translation{\psi}
    =
    \psi\big[\varphi/\auxatom{\varphi}\mid\varphi\in\negocc{\psi}\big]
    \wedge
    \bigwedge_{\varphi\in\negocc{\psi}}(\neg\auxatom{\varphi}\vee\auxatom{\varphi})
    \wedge
    \bigwedge_{\varphi\in\negocc{\psi}}\neg \neg (\varphi\leftrightarrow\auxatom{\varphi})
    \ .
  \]
\end{definition}
% --------------------------------------------------------------------------------
%
% --------------------------------------------------------------------------------
\begin{example}\label{ex:lp:translation}
  Let $\psi$ be $\neg a\to b\vee\neg \neg(c\wedge\neg d)$.
  Then,
  \begin{align*}
    \negocc{\psi}      =\ & \{\neg a,\neg\neg(c\wedge\neg d)\}\\
    \translation{\psi} =\ & (\auxatom{\neg a}\to b\vee\auxatom{\neg\neg (c\wedge\neg d)})\;        \wedge\\
                          & (\auxatom{\neg a}\vee\neg\auxatom{\neg a})                     \wedge
                            (\auxatom{\neg\neg(c\wedge\neg d)}\vee\neg\auxatom{\neg\neg (c\wedge\neg d)})       \\
                          & \neg\neg({\neg a}         \leftrightarrow\auxatom{\neg a})     \wedge
                            \neg\neg({\neg\neg(c\wedge\neg d)}\leftrightarrow\auxatom{\neg\neg(c\wedge\neg d)})
  \end{align*}
\end{example}
% --------------------------------------------------------------------------------

With the translation from Definition~\ref{def:translation}, we obtain an analogous conservative extension result as above.
%
% --------------------------------------------------------------------------------
\begin{theorem}\label{thm:extension:definition:negated}
  Let $\psi$ be a formula over \At.

  Then, we have
  \(
  \psi \cong_{\At}\translation{\psi}
  \).
\end{theorem}
% --------------------------------------------------------------------------------
%
In analogy to Proposition~\ref{prop:extension:definition:one:one},
we get a one-to-one correspondence between the stable models of both formulas.
%
% --------------------------------------------------------------------------------
\begin{theorem}\label{thm:extension:definition:negated:one:one}
  Let $\psi$ be a formula over \At.

  \begin{enumerate}
  \item If $T\subseteq\At$ is a stable model of $\psi$,
    then $T \cup \{\auxatom{\varphi} \mid \varphi \in \negocc{\psi} \text{ and }T \models \varphi \}$
    is a stable model of $\translation{\psi}$.
  \item If $T\subseteq (\At \cup \{\auxatom{\varphi} \mid \varphi \in \negocc{\psi}\})$ is a stable model of $\translation{\psi}$,
    then $T\cap\At$ is a stable model of $\psi$.
  \end{enumerate}
\end{theorem}
% --------------------------------------------------------------------------------
%
% --------------------------------------------------------------------------------
For instance, $\{b\}$ is a stable model of the formula $\psi=\neg a\to b\vee\neg \neg(c\wedge\neg d)$ from Example~\ref{ex:lp:translation}.
From Theorem~\ref{thm:extension:definition:negated}, $\{\auxatom{\neg a}, b\}$ is a stable model of $\translation{\psi}$.
Conversely, from the stable model $\{\auxatom{\neg a}, b\}$ of $\translation{\psi}$,
we get the stable model $\{b\}$ of $\psi$ by dropping the new atoms.
% --------------------------------------------------------------------------------

%%% Local Variables:
%%% mode: latex
%%% TeX-master: "paper"
%%% End:

\section{Austere Answer Set Programming}

In this section,
we restrict the application of our formula translation to logic programs.
Although we focus on normal programs,
a similar development with other classes of logic programs, like disjunctive ones, can be done accordingly.

For simplicity, we write $\auxatm{a}$ instead of $\auxatom{\neg a}$ for $a\in \At$ and let $\{\auxatm{a}\}$ stand for $\auxatm{a}\vee\neg\auxatm{a}$.
Note that, for a rule $r$ as in \eqref{rule}, the set $\negocc{r}$ consists of negative literals only.
The next two definitions specialize our translation of formulas to logic programs.
%
% --------------------------------------------------------------------------------
\begin{definition}\label{def:rule:translation}
  Let $r$ be a rule over \At\ as in \eqref{rule} with $m\geq 1$.

  Then, we define
  \begin{align*}
    \translation{r}
    &\textstyle =
      r\big[\neg a/\auxatm{a}\mid \neg a \in \negocc{r}]
      \cup
      \bigcup_{\neg a \in \negocc{r}}
      \left\{\{\auxatm{a}\} \leftarrow\right\}
      \cup
      \bigcup_{\neg a \in \negocc{r}}
      \left\{
      \begin{array}{c@{\ }l}
        \leftarrow&       a\wedge     \auxatm{a} \\
        \leftarrow&  \neg a\wedge \neg\auxatm{a}
      \end{array}
      \right\}
  \end{align*}
\end{definition}
% --------------------------------------------------------------------------------
%
% --------------------------------------------------------------------------------
\begin{definition}\label{def:program:translation}
  Let $P$ be a logic program over \At.
  Then,
  \(
  \translation{P} = \bigcup_{r \in P}	\translation{r}
  \).
\end{definition}
% --------------------------------------------------------------------------------
%
This translation substitutes negated literals in rule bodies with fresh atoms and
adds a choice rule along with a pair of integrity constraints providing an equivalence between
the eliminated negated body literals and the substituted atoms.

By applying the above results in the setting of logic programs,
we get that
a logic program and its translation have the same stable models when restricted to the original vocabulary.
%
% --------------------------------------------------------------------------------
\begin{corollary}\label{cor:extension:definition:program}
  Let $P$ be a logic program over \At.

  Then, we have
  \(
  P \cong_{\At} \translation{P}
  \)
\end{corollary}
% --------------------------------------------------------------------------------
%
In other words,
every stable model of a logic program can be extended to a stable model of its translation
and vice versa.
%
% --------------------------------------------------------------------------------
\begin{corollary}\label{cor:extension:definition:program:one:one}
  Let $P$ be a logic program over \At.
  \begin{enumerate}
  \item If $T\subseteq\At$ is a stable model of $P$,
    then $T \cup \{\auxatm{a} \mid \neg a \in \negocc{P} \text{ and } a \not\in T \}$ is a stable model of \translation{P}.
  \item $T\subseteq(\At\cup \{\auxatm{a} \mid \neg a \in \negocc{P} \}$  is a stable model of \translation{P},
    then $T\cap\At$ is a stable model of $P$.
  \end{enumerate}
\end{corollary}
% --------------------------------------------------------------------------------

For illustration,
consider the following example.
%
% --------------------------------------------------------------------------------
\begin{example}\label{ex:program:translation}
  Consider the normal logic program $P$:
  \[
    \begin{array}{rcl}
      a &\leftarrow& \\
      b &\leftarrow& \neg c\\
      c &\leftarrow& \neg b\\
      d &\leftarrow& a \wedge \neg c
    \end{array}
  \]
  Then, \translation{P} is:
  \[
    \begin{array}{rclrclrcl}
      a &\leftarrow&                     & \{\auxatm{b} \} &\leftarrow&                               & \{\auxatm{c} \} &\leftarrow& \\
      b &\leftarrow& \auxatm{c}          &                 &\leftarrow&  b\wedge\auxatm{b}            &                 &\leftarrow& c\wedge\auxatm{c}\\
      c &\leftarrow& \auxatm{b}          &                 &\leftarrow&  \neg b\wedge \neg \auxatm{b} &                 &\leftarrow& \neg c\wedge \neg \auxatm{c} \\
      d &\leftarrow& a \wedge \auxatm{c}
    \end{array}
  \]
  The stable models of $P$ are $\{a,b,d\}$ and $\{a,c\}$ and the ones of \translation{P} are $\{a,b,d,\auxatm{c}\}$ and $\{a,c,\auxatm{b}\}$, respectively.
\end{example}
The example underlines that our translation maps normal rules to definite ones along with choices and pairs of integrity constraints.
In other words,
it can be seen as a means for expressing normal logic programs in the form of programs with facts, definite rules, choice rules and integrity
constraints over an extended vocabulary.
We call this class of programs \emph{austere logic programs}, and further elaborate upon them in the following.

\subsection{Austere Logic Programs}

We define austere logic programs according to the decomposition put forward in the introduction.
%
% --------------------------------------------------------------------------------
\begin{definition}[Austere logic program]\label{def:austere:program}
  An \emph{austere logic program} is a quadruple \linebreak
  \(
  (F,C,D,I)
  \)
  consisting of
  a set $F$ of facts,
  a set $C$ of choices,\footnote{That is, choice rules without body literals.}
  a set $D$ of definite rules, and
  a set $I$ of integrity constraints.
\end{definition}
% --------------------------------------------------------------------------------
%
A set of atoms is a stable model of an austere logic program, if it is a stable model of the union of all four components.

In view of the above results, austere logic programs can be regarded as a normalform for normal logic programs.
\comment{JR: This sounds a bit strange. Typically a normal form of some language is a subset of that language.
But here the austere programs have choice rules, that do not belong to normal logic programs.
In other words, the normal form of a normal logic program is not a normal logic program.\par
We could mention that it is useful for theory and practice to resort to such a form.
\par T: Right, we may have to talk about "even loops", sigh!
\par J: we must be careful then when we move to stratified negation, because there we forbid these loops.\par
Another option is simply to say that austere programs provide a normal form for normal logic programs with choice rules, or something like that.}
%
% --------------------------------------------------------------------------------
\begin{corollary}\label{cor:austere:normalform}
  Every normal logic program can be expressed as an austere logic program and vice versa.
\end{corollary}
% --------------------------------------------------------------------------------
%
The converse follows from the fact that choice rules are expressible by a pair of normal rules~\cite{siniso02a}.

In fact, the (instantiation of) Listing~\ref{lp:ham} constitutes an austere logic program.
To see this observe that
\begin{itemize}
\item
  lines~1-3 provide facts, $F$, capturing the problem instance,
  here giving the specification of a graph;
\item
  Line~5 provides choices, $C$, whose instantiation is derived from facts in the previous lines.
  Grounding expands this rule to several plain choice rules with empty bodies;
\item
  lines 5-6 list definite rules, $D$, defining (auxiliary) predicates used in the integrity constraints;
\item
  finally, integrity constraints, $I$, are given in lines 7-9, stating conditions that solutions must satisfy.
\end{itemize}

This example nicely illustrates a distinguishing feature of austere logic programs,
namely, the \emph{compartmentalization} of the program parts underlying ASP's guess-define-check encoding methodology
(along with its strict separation of instance and encoding):
The problem instance is described by means of
\begin{itemize}
\item the facts in $F$
\end{itemize}
and the problem encoding confines
\begin{itemize}
\item non-deterministic choices to $C$,
\item the deterministic extension of the taken decisions to $D$, and
\item the test of the obtained extension to $I$.
\end{itemize}
This separation also confines the sources of multiple or non-existing stable models to well-defined locations, namely, $C$ and $I$, respectively
(rather than spreading them over several circular rules; see below).
As well,
the rather restricted syntax of each compartment gives rise to a very simple operational semantics of austere logic programs,
as we see in the next section.

%%% Local Variables:
%%% mode: latex
%%% TeX-master: "paper"
%%% End:
\subsection{Operational Semantics}

In our experience,
a major factor behind the popularity of the approach sketched in the introductory section
lies in the possibility to intuitively form stable models along the order of the rules in a program.
In fact,
the simple nature of austere logic programs provides a straightforward scheme for computing stable models
by means of the well-known immediate consequence operator,
whose iteration mimics this proceeding.
Moreover,
the simplicity of the computation provides the first evidence of the value of austere logic programs as a normalform.

The operational semantics of austere logic programs follows ASP's \emph{guess-define-check} methodology.
In fact,
the only non-determinism in austere logic programs is comprised of choice rules.
Hence,
once choices are made,
we may adapt well-known deterministic bottom-up computation techniques for computing stable models.
However,
the results of this construction provide merely candidate solutions
that still need to satisfy all integrity constraints.
If this succeeds, they constitute stable models of the austere program.

Let us make this precise for an austere logic program $(F,C,D,I)$ in what follows.
To make choices and inject them into the bottom-up computation,
we translate the entire set of choices, $C$, into a set of facts:
\[
  F_C = \{{a \leftarrow}\mid{\{a\} \leftarrow }\in C\}
\]
A subset of $F_C$, the original facts $F$, along with the definite program $D$ are then passed to a
corresponding consequence operator that determines a unique stable model candidate.
More precisely,
the $T_P$ operator of a definite program $P$ is defined for an interpretation $X$ as follows~\cite{lloyd87}:
\[
T_P(X) = \{ l_1 \mid (l_1\leftarrow l_{m+1}\wedge\dots\wedge l_n) \in P,\ X \models l_{m+1}\wedge\dots\wedge l_n \}
\]
With this, the candidate solutions of an austere program can be defined.
%
% --------------------------------------------------------------------------------
\begin{definition}\label{def:candidate:stable:model}
  Let $(F,C,D,I)$ be an austere logic program over \At.

  We define a set $X\subseteq\At$ of atoms as a candidate stable model of $(F,C,D,I)$,
  if $X$ is the least fixpoint of $T_{F\cup C'\cup D}$ for some $C' \subseteq F_C$.
\end{definition}
% --------------------------------------------------------------------------------
%
The existence of the least fixpoint is warranted by the monotonicity of $T_{F\cup C'\cup D}$~\cite{lloyd87}.
Similar to traditional ASP,
several candidate models are obtained via the different choices of $C'$.

While the choice of $C'$ constitutes the \emph{guess} part and the definite rules in $D$ the \emph{define} part of the approach,
the \emph{check} part is accomplished by the integrity constraints in~$I$.
%
% --------------------------------------------------------------------------------
\begin{proposition}\label{prop:austere:stable:models}
  Let $(F,C,D,I)$ be an austere logic program over \At\
  and $X\subseteq\At$.

  Then, $X$ is a stable model of $(F,C,D,I)$
  iff
  $X$ is a candidate stable model of $(F,C,D,I)$ such that $X \models I$.
\end{proposition}
% --------------------------------------------------------------------------------

We illustrate the computation of stable models of austere logic programs in the following example.
\begin{example}
  Consider the austere logic program $P$
  \[
    \begin{array}{rcl}
      a &\leftarrow& \\
      \{b\} &\leftarrow& \\
      c &\leftarrow& b\\
        &\leftarrow& a \wedge \neg c\\
    \end{array}
  \]
  We get the candidate stable models $\{a,b,c\}$ and $\{a\}$ from the first three rules depending on whether
  we choose $b$ to be true or not,
  that is, whether we add the fact $b \leftarrow$ or not.
  Then,
  on testing them against the integrity constraint expressed by the fourth rule,
  we see that $\{a,b,c\}$ is indeed a stable model, since it satisfies the integrity constraint,
  while set~$\{a\}$ is not a stable model since checking the integrity constraint fails.
\end{example}

A major intention of austere logic programs is to confine the actual guess and check of an encoding to dedicated components,
namely, the choices in $C$ and constraints in $I$.
The definite rules in $D$ help us to analyze and/or extend the solution candidate induced by the facts $F$ and the actual choices in $C'$.
The emerging candidate is then evaluated by the integrity constraints in $I$.
This stresses once more the idea that the extension of a guessed solution candidate should be deterministic;
it elaborates the guess but refrains from introducing any ambiguities.
This is guaranteed by the definite rules used in austere programs.
%
% --------------------------------------------------------------------------------
\begin{observation}\label{obs:uniqueness}
  For any austere logic program $(F,C,D,I)$ and $C' \subseteq F_C$,
  the logic program $F\cup C'\cup D$ has a unique stable model.
\end{observation}
% --------------------------------------------------------------------------------
%
This principle is also in accord with~\cite{niemela08b}, where stratified logic programs are used instead of definite ones (see below).

%%% Local Variables:
%%% mode: latex
%%% TeX-master: "paper"
%%% End:

\section{Easy Answer Set Programming}

Austere logic programs provide a greatly simplified format that reflects ASP's \emph{guess-define-check} methodology~\cite{lifschitz02a}
for writing encodings.
Their simple structure allows for translating the methodology into
an intuitive process that consists of
making non-deterministic choices, followed by a deterministic bottom-up computation, and a final consistency check.

In what follows,
we want to turn the underlying principles into a modeling methodology for ASP that aims at laypersons.
To this end, we leave the propositional setting and aim at full-featured input languages of ASP systems like \clingo~\cite{gekakasc17a} and \dlv~\cite{dlv03a}.
Accordingly, we shift our attention to predicate symbols rather than propositions and let the terms `logic program', `rule', etc.\ refer to these languages
without providing a technical account (cf.~\cite{gehakalisc15a,cafageiakakrlemarisc19a}).
Moreover, we allow for normal rules instead of definite ones as well as aggregate literals in bodies
in order to accommodate the richness of existing ASP modeling languages.

The admission of normal rules comes at the expense of losing control over the origin of multiple or non-existing stable models as well as
over a deterministic development of guessed solutions.
In fact,
the idea of \emph{Easy Answer Set Programming} (ezASP) is to pursue the principles underlying austere logic programs
without enforcing them through a severely restricted syntax.
However,
rather than having the user fully absorb the loss in control,
we shift our focus to a well-founded development of ASP encodings,
according to which predicates are defined on top of previously defined predicates
(or facts).
This parallels the structure and the resulting operational semantics of austere logic programs.

To this end,
we start by capturing dependencies among predicates~\cite{apblwa87a}.
%
% --------------------------------------------------------------------------------
\begin{definition}
  Let $P$ be a logic program.
  \begin{itemize}
  \item A predicate symbol $p$ \emph{depends} upon a predicate symbol $q$,
    if there is a rule in $P$ with $p$ on its left-hand side and $q$ on its right-hand side.

    If $p$ depends on $q$ and $q$ depends on $r$, then $p$ depends on $r$, too.
  \item The \emph{definition} of a predicate symbol $p$ is the subset of $P$
    consisting of all rules with $p$ on their left-hand side.
  \end{itemize}
\end{definition}
% --------------------------------------------------------------------------------
%
We denote the definition of a predicate symbol $p$ in $P$ by $\mathit{def}(p)$
and view integrity constraints as rules defining $\bot$.

Our next definition makes precise what we mean by a well-founded development of a logic program.
\footnote{The term \emph{stratification} differs from the one used in the literature~\cite{apblwa87a}.}
%
% --------------------------------------------------------------------------------
\begin{definition}\label{def:stratification}
  Let $P$ be a logic program.

  We define a partition
  \(
  (P_1,\dots, P_n)
  \)
  of $P$ as a \emph{stratification} of $P$,
  if
  \begin{enumerate}
  \item $\mathit{def}(p)\subseteq P_i$ for all predicate symbols $p$ and some $i\in\{1,\dots,n\}$
    and
  \item if
    $p$ depends on $q$,
    $\mathit{def}(p)\subseteq P_i$,
    and
    $\mathit{def}(q)\subseteq P_j$    for some $i,j\in\{1,\dots,n\}$,
    then
    \begin{enumerate}
    \item $i>j$ unless $q$ depends on $p$, and
    \item $i=j$ otherwise
    \end{enumerate}
  \end{enumerate}
\end{definition}
% --------------------------------------------------------------------------------
%
Any normal logic program has such a stratification.
One way to see this is that mutually recursive programs can be trivially stratified via a single partition.
For instance, this applies to both programs
\(
\{a\leftarrow b, b\leftarrow a\}
\)
and
\(
\{a\leftarrow \neg b, b\leftarrow \neg a\}
\)
in which $a$ and $b$ mutually depend upon each other.
Accordingly, similar recursive structures in larger programs are confined to single partitions, as required by (\emph{2b}) above.

With it, we are ready to give shape to the concept of an \emph{easy logic program}.
%
% --------------------------------------------------------------------------------
\begin{definition}[Easy logic program]\label{def:easy:program}
  An easy logic program is a logic program
  having stratification
  \(
  (F,C,D_1,\dots,D_n,I)
  \)
  such that
  $F$ is a set of facts,
  $C$ is a set of choice rules,
  $D_i$ is a set of normal rules for $i=1,\dots,n$, and
  $I$ is a set of integrity constraints.
\end{definition}
% --------------------------------------------------------------------------------
\comment{JR: I don't understand the definition: normal logic programs have no choice rules.
We say that they can be expressed by normal rules, but
strictly speaking choice rules are not normal rules, and
they require non-stratified negation (that we eliminate later) to be expressed
by normal programs.\par
T: partly fixed}
As in traditional ASP, we often divide a logic program into facts representing a problem instance and the actual encoding of the problem class.
For easy programs, this amounts to separating $F$ from $(C,D_1,\dots,D_n,I)$.

Clearly, an austere logic program is also an easy one.

Thus, the program in Listing~\ref{lp:ham} is also an easy logic program having the stratification
\[
(\{\mathrm{1},\mathrm{2},\mathrm{3}\},\{\mathrm{5}\},\{\mathrm{6},\mathrm{7}\},\{\mathrm{8},\mathrm{9},\mathrm{10}\})
\]
where each number stands for the rules in the respective line.

Predicates \texttt{node/1}, \texttt{edge/2}, and \texttt{start/1} are only used to form facts or occur in rule bodies.
Hence, they do not depend on any other predicates and can be put together in a single component, $F$.
This makes sense since they usually constitute the problem instance.
Putting them first reflects that the predicates in the actual encoding usually refer to them.
The choices in $C$ provide a solution candidate that is checked by means of the rules in the following components.
In our case, the guessed extension of predicate \texttt{hc/2} in Line~5 is simply a subset of all edges provided by predicate \texttt{edge/2}.
Tests for being a path or even a cycle are postponed to the \emph{define-check} part:
The rules in $\{\mathrm{6},\mathrm{7}\}$, that is, $D_1$, define the auxiliary predicate \texttt{reached/1},
and aim at analyzing and/or extending our guessed solution candidate by
telling us which nodes are reachable via the instances of \texttt{hc/2} from the start node.
The actual checks are then conducted by the integrity constraints, $I$, in the final partition $\{\mathrm{8},\mathrm{9},\mathrm{10}\}$.
At this point, the solution candidate along with all auxiliary atoms are derived and ready to be checked.
Line~8 tests whether each node is reached in the solution at hand,
while lines~9 and~10 make sure that a valid cycle never enters or leaves any node twice.

Finally, it is instructive to verify that strata $\{\mathrm{5}\}$ and $\{\mathrm{6},\mathrm{7}\}$ cannot be reversed or merged.
We observe that
\begin{itemize}
\item \texttt{hc/2} depends on \texttt{edge/2} only,
\end{itemize}
while
\begin{itemize}
\item \texttt{reached/1} depends on \texttt{hc/2}, \texttt{edge/2}, \texttt{start/1}, and itself,
\end{itemize}
and no other dependencies.
The rules defining \texttt{hc/2} and \texttt{reached/1} must belong to the same partition, respectively, as required by (\emph{2a}) above.
Thus,
$\{\mathrm{5}\}\subseteq P_i$
and
$\{\mathrm{6},\mathrm{7}\}\subseteq P_j$
for some $i,j$.
Because \texttt{reached/1} depends on \texttt{hc/2} and not vice versa, we get $i<j$.
This dependency rules out an inverse arrangement,
and the fact that it is not symmetric excludes a joint membership of both definitions in the same partition,
as stipulated by (\emph{2b}) above.

\subsection{Modeling Methodology}

The backbone of easy ASP's modeling methodology is the structure imposed on its programs in Definition~\ref{def:easy:program}.
This allows us to encode problems by building concepts on top of each other.
Also, its structure allows for staying in full tune with ASP's \emph{guess-define-check} methodology~\cite{lifschitz02a}
by offering well-defined spots for all three parts.

Easy logic programs tolerate normal rules
in order to encompass full-featured ASP modeling languages.
Consequently, the interplay of the guess, define, and check parts of an easy logic program defies any control.
To tame this opening, we propose to carry over Observation~\ref{obs:uniqueness} to easy logic programs:
For any easy logic program
\(
(F,C,D_1,\dots,D_n,I)
\)
and $C' \subseteq F_C$,
the logic program $F\cup C'\cup D_1\cup\dots\cup D_n$ should have a unique stable model.
Even better if this can be obtained in a deterministic way.

This leads us to the following advice on easy ASP modeling:
\begin{enumerate}
\item Compartmentalize a logic program into
  facts, $F$,
  choice rules, $C$,
  normal rules, $D_1\cup\dots\cup D_n$, and
  integrity constraints $I$,

  such that the overall logic program has stratification $(F,C,D_1,\dots,D_n,I)$.
\item
  Aim at defining one predicate per stratum $D_i$
  and
  avoid cycles within each $D_i$ for $i=1,\dots,n$.

\item Ensure that $F\cup C'\cup D_1\cup\dots\cup D_n$ has a unique stable model for any $C' \subseteq F_C$.
\end{enumerate}
While the first two conditions have a syntactic nature and can thus be checked automatically,
the last one refers to semantics
and, to the best of our knowledge, has only sufficient but no known necessary syntactic counterparts.
One is to restrict $D_1\cup\dots\cup D_n$ to definite rules as in austere programs,
the other is to use stratified negation, as proposed in~\cite{niemela08b} and detailed in the next section.

Our favorite is to stipulate that
\(
{F\cup C'\cup D_1\cup\dots\cup D_n}$ has a total well-founded model~\cite{gerosc91a} for any $C' \subseteq F_C
\)
but unfortunately, we are unaware of any syntactic class of logic programs warranting this condition beyond the ones mentioned above.

%%% Local Variables:
%%% mode: latex
%%% TeX-master: "paper"
%%% End:

\subsection{Stratified Negation}
\label{sec:stratified:negation}

The purpose of stratified negation is to eliminate the (problematic) case of recursion through negation.
What makes this type of recursion problematic is that it may eliminate stable models and that the actual source may be spread over several rules.
To give some concise examples, consider the programs
\(
\{a\leftarrow \neg a\}
\)
and
\(
\{
a\leftarrow \neg b,
b\leftarrow \neg c,
c\leftarrow \neg a
\}
\)
admitting no stable models.
Following the dependencies in both examples,
we count one and three dependencies, respectively, all of which pass through negated body literals.
More generally, cyclic dependencies traversing an odd number of negated body literals (not necessarily consecutively) are known sources of incoherence.
Conversely, an even number of such occurrences on a cycle is not harmful but spawns alternatives, usually manifested in multiple stable models.
To see this, consider the program
\(
{\{a\leftarrow \neg b, b\leftarrow \neg a\}}
\)
producing two stable models.
Neither type of rule interaction is admitted in austere logic programs.
Rather the idea is to confine the sources of multiple and eliminated stable models to dedicated components,
namely, choices and integrity constraints.
The same idea was put forward by Niemel{\"{a}} in~\cite{niemela08b}
yet by admitting a more general setting than definite rules
by advocating the concept of stratified negation.

To eliminate the discussed cyclic constellations,
stratified negation imposes an additional constraint on the stratification of a logic program:
Given the prerequisites of Definition~\ref{def:stratification}, we define:
\begin{enumerate}\setcounter{enumi}{2}
\item If a predicate symbol $q$ occurs in a negative body literal of a rule in $P_i$,
  then $\mathit{def}(q)\subseteq P_j$ for some $j<i$.
\end{enumerate}
In other words,
while the definitions of predicates appearing positively in rule bodies may appear in a lower or equal partition,
the ones of negatively occurring predicates are restricted to lower components.
Although this condition tolerates positive recursion as in
\(
\{a\leftarrow b, b\leftarrow a\}
\),
it rules out negative recursion as in the above programs.
Since using programs with stratified negation rather than definite programs generalizes austere logic programs,
their combination with choices and integrity constraints is also as expressive as full ASP~\cite{niemela08b}.

An example of stratified negation can be found in Listing~\ref{lst:toh}.
The negative literal in Line~5 refers to a predicate defined --- beforehand --- in Line~8.

An attractive feature of normal logic programs with stratified negation is that they yield a unique stable model,
just as with austere programs (cf.~Observation~\ref{obs:uniqueness}).
Hence, they provide an interesting generalization of definite rules
maintaining the property of deterministically extending guessed solution candidates.

%%% Local Variables:
%%% mode: latex
%%% TeX-master: "paper"
%%% End:

\subsection{Complex Constraints}
\label{sec:aggregates}

As mentioned, we aim at accommodating complex language constructs as aggregates in order to leverage the full expressiveness of ASP's
modeling languages.
For instance, we may replace lines~9 and~10 in Listing~\ref{lp:ham} by
\begin{lstlisting}[firstnumber=9]
:- { hc(U,V) } >= 2, node(U).
:- { hc(U,V) } >= 2, node(V).
\end{lstlisting}
without violating its stratification.

More generally,
a rule with an aggregate
`\(
\#\mathit{op}\{l_1,\dots, l_m\}\prec k
\)'
in the consequent can be represented with choice rules along with an integrity constraint,
as shown in~\cite{siniso02a}.
That is, we can replace any rule of form
\[
  \#\mathit{op}\{l_1,\dots, l_m\}\prec k\leftarrow l_{m+1}\wedge\dots\wedge l_n
\]
by\footnote{In practice, a set of such choice rules can be represented by a single one of form
$\{l_1,\dots,l_m\}\leftarrow l_{m+1}\wedge\dots\wedge l_n$.}
\begin{align*}
  \{l_i\}&\leftarrow l_{m+1}\wedge\dots\wedge l_n         &\text{ for }i=1,\dots,m\text{ and}\\
  \bot&\leftarrow \neg(\#\mathit{op}\{l_1,\dots, l_m\}\prec k)\wedge l_{m+1}\wedge\dots\wedge l_n \ .&
\end{align*}
This allows us to integrate aggregate literals into easy logic programs without sacrificing expressiveness.

In fact, many encodings build upon restricted choices that are easily eliminated by such a transformation.
A very simple example is graph coloring.
Assume a problem instance is given in terms of facts
\lstinline{node/1},
\lstinline{edge/2}, and
\lstinline{color/1}.
A corresponding encoding is given by the following two rules:
% --------------------------------------------------------------------------------
\begin{lstlisting}
{ assign(X,C) : color(C)} = 1 :- node(X).
:- edge(X,Y), assign(X,C), assign(Y,C).
\end{lstlisting}
% --------------------------------------------------------------------------------
Note that the aggregate in the consequent of Line~1 is a shortcut for a \textit{\#count} aggregate.

To eliminate the restricted choice from the consequent in Line~1,
we may apply the above transformation to obtain the following easy encoding:
% --------------------------------------------------------------------------------
\begin{lstlisting}
{ assign(X,C) } :- node(X), color(C).
:- not { assign(X,C) : color(C)} = 1, node(X).
:- edge(X,Y), assign(X,C), assign(Y,C).
\end{lstlisting}
% --------------------------------------------------------------------------------
%
Given some set of facts, $F$, this encoding amounts to the easy logic programs \linebreak
\(
(F,\{\mathrm{1}\},\{\mathrm{2}\},\{\mathrm{3}\}).
\)

The decomposition into a choice and its restriction may appear unnecessary to the experienced ASP modeler.
However, we feel that such a separation adds clarity and is preferable to language constructs combining several aspects,
at least for ASP beginners.
Also, it may be worth noting that this decomposition is done anyway by an ASP system and hence brings about no performance loss.

Two further examples of easy logic programs are given in Listing~\ref{lst:queens} and~\ref{lst:toh},
solving the Queens and the Tower-of-Hanoi puzzles both with parameter~\texttt{n}.
\footnote{This parameter is either added from the command line via option \texttt{--const} or a default added via directive \texttt{\#const}
  (see~\cite{PotasscoUserGuide} for details).}
%
% --------------------------------------------------------------------------------
\begin{lstlisting}[float=ht,caption={An easy logic program for the $\mathtt{n}$-Queens puzzle},label=lst:queens]
{ queen(1..n,1..n) }.

d1(I,J,I-J+n) :- I = 1..n, J = 1..n.
d2(I,J,I+J-1) :- I = 1..n, J = 1..n.

:- { queen(I,1..n) } != 1, I = 1..n.
:- { queen(1..n,J) } != 1, J = 1..n.

:- { queen(I,J) : d1(I,J,D) } > 1, D=1..n*2-1.
:- { queen(I,J) : d2(I,J,D) } > 1, D=1..n*2-1.
\end{lstlisting}
% --------------------------------------------------------------------------------
%
% --------------------------------------------------------------------------------
\begin{lstlisting}[float=ht,caption={An easy logic program for a Towers-of-Hanoi puzzle (for plans of length $\mathtt{n}$)},label=lst:toh]
peg(a;b;c).
disk(1..4).
init_on(1..4,a).
goal_on(1..4,c).

{ move(D,P,T) : disk(D), peg(P) } :- T = 1..n.

move(D,T)          :- move(D,_,T).

on(D,P,0)          :- init_on(D,P).
on(D,P,T)          :- move(D,P,T).
on(D,P,T+1)        :- on(D,P,T), not move(D,T+1), T < n.

blocked(D-1,P,T+1) :- on(D,P,T), T < n.
blocked(D-1,P,T)   :- blocked(D,P,T), disk(D).

:- { move(D,P,T) : disk(D), peg(P) } != 1, T = 1..n.

:- move(D,P,T), blocked(D-1,P,T).
:- move(D,T), on(D,P,T-1), blocked(D,P,T).
:- not 1 { on(D,P,T) } 1, disk(D), T = 1..n.

:- goal_on(D,P), not on(D,P,n).
\end{lstlisting}
% --------------------------------------------------------------------------------
%
While the easy logic program for the $\mathtt{n}$-Queens puzzle has the format
\[
  (\emptyset,\{\mathrm{1}\},\{\mathrm{3,4}\},\{\mathrm{6,7}\},\{\mathrm{9,10}\}),
\]
the one for the Tower-of-Hanoi puzzle can be partitioned into
\[
  (\{\mathrm{1,2,3,4}\},\{\mathrm{6}\},\{\mathrm{8}\},\{\mathrm{10,11,12}\},\{\mathrm{14,15}\},\{\mathrm{17,19,20,21,23}\})
  \ .
\]

%%% Local Variables:
%%% mode: latex
%%% TeX-master: "paper"
%%% End:

% \input{optimization}
% \input{debugging}
\subsection{Limitations}
\label{sec:limitations}

The methodology of ezASP has its limits.
For instance, sometimes it is convenient to make choices depending on previous choices.
Examples of this are the combination of routing and scheduling, as in train scheduling~\cite{abjoossctowa19a},
or the formalization of frame axioms in (multi-valued) planning advocated in~\cite{leliya13a}.
Another type of encodings escaping our methodology occurs in meta programming,
in which usually a single predicate, like \texttt{holds}, is used and atoms are represented as its arguments.
Thus, for applying the ezASP methodology, one had to refine the concept of stratification to access the term level
in order to capture the underlying structure of the program.
And finally, formalizations of planning and reasoning about actions involve
the formalization of effect and inertia laws that are usually self-referential on the predicate level
(sometimes resolved on the term level, through situation terms or time stamps).
A typical example of circular inertia laws is the following:
\begin{lstlisting}[numbers=none]
 holds(F,T) :-  holds(F,T-1), not -holds(F,T).
-holds(F,T) :- -holds(F,T-1), not  holds(F,T).
\end{lstlisting}
Here, `\texttt{-}' denotes classical negation, and \texttt{F} and \texttt{T} stand for (reified) atoms and time points.
On the other hand, the sophistication of the given examples illustrates that they are usually not addressed by beginners but rather experts in
ASP for which the strict adherence to ezASP is less necessary.

%%% Local Variables:
%%% mode: latex
%%% TeX-master: "paper"
%%% End:

%%% Local Variables:
%%% mode: latex
%%% TeX-master: "paper"
%%% End:

\section{Related Work}
\label{sec:related}

Apart from advocating the idea illustrated in the introduction,
Ilkka Niemel{\"{a}} also showed in~\cite{niemela08b} that
negative body literals can be replaced by a new atom for which
a choice needs to be made whether to include it in the model or not;
and
such that a model cannot contain both the new atom and the atom of the replaced literal
but one of them needs to be included.
This technique amounts exactly to the transformation in Definition~\ref{def:rule:translation}
and traces back to Abductive logic programming~\cite{eshkow89a,denkak02a}.
Indeed, it was already shown in \cite{grsaza01a} that for DATALOG queries the expressive power of stable model semantics can be achieved via stratified negation and choices.

We elaborated upon this idea in several ways.
First, we have shown that the full expressiveness of normal logic programs can even be achieved with definite rules rather than normal ones with
stratified negation.
Second, we have provided a strong equivalence result that allows for applying the transformation in Definition~\ref{def:rule:translation} to selected
rules only.
Third, we have generalized the idea by means of the logic of Here-and-There, which made it applicable to other fragments of logic programs.
And finally, this investigation has revealed that the roots of the idea lie in the logical principle of extension by definition.

Over the last decades many more related ideas were presented in the literature.
For instance, in \cite{felomisu93a}, normal logic programs are translated into positive disjunctive programs by introducing new atoms for negative literals.
Also, strong negation is usually compiled away via the introduction of new atoms along with integrity constraints excluding that both the original atom
and the atom representing its strong negation hold~\cite{gellif90a}.
The principle of extension by definition was also used in~\cite{ferlif02a} to prove properties about programs with nested expressions.
EzASP is closely related to the paradigm of IDP~\cite{debobrjade18a},
where the program parts $F$, $C$ and $I$ are expressed in first-order logic,
while the $D_i$'s form inductive definitions.
Finally, in~\cite{delitrve19a},
the authors propose an informal semantics for logic programs based on the guess-define-check methodology,
that are similar to the easy logic programs that we introduce in this paper.

%%% Local Variables:
%%% mode: latex
%%% TeX-master: "paper"
%%% End:

\section{Conclusion}\label{sec:discussion}

We have revisited an old idea from the literature on logic programming under stable model semantics
and elaborated upon it in several ways.
We started by tracing it back to the principle of extension by definition.
The resulting formalization in the setting of the logic of Here-and-there provides us with a logical framework
that can be instantiated in various ways.
Along these lines,
we have shown that normal logic programs can be reduced to choices, definite rules, and integrity constraints,
while keeping the same expressiveness as the original program.
A major advantage of this austere format is that it confines non-determinism and incoherence to well-defined spots in the program.
The resulting class of austere logic programs could play a similar role in ASP as formulas in conjunctive normal form in classical logic.

Drawing on the properties observed on austere logic program,
we put forward the modeling methodology of ezASP.
The idea is to compensate for the lacking guarantees provided by the restricted format of austere programs
by following a sequential structure when expressing a problem in terms of a logic program.
This makes use of the well-known concept of stratification to refine ASP's traditional guess-define-check methodology.
Although the ordering of rules may seem to disagree with the holy grail of full declarativeness,
we evidence its great value in introducing beginners to ASP.
Also, many encodings by experienced ASP users follow the very same pattern.

Moreover, the ezASP paradigm aligns very well with that of \emph{achievements}~\cite{lifschitz17a}
that aims not only at easily understandable but moreover provably correct programs.
To this end, formal properties are asserted in between a listing of rules to express what has been achieved up to that point.
Extending ezASP with achievements and automatically guiding the program development with ASP verifiers, like \anthem~\cite{lilusc19a},
appears to us as a highly interesting avenue of future research.
In this context, it will also be interesting to consider the components of an easy logic program as modules with an attached input-output
specification, so that the meaning of the overall program emerges from the composition of all components.
This would allow for successive refinements of programs' components, while maintaining their specification.

%%% Local Variables:
%%% mode: latex
%%% TeX-master: "paper"
%%% End:

\bibliographystyle{splncs04}
\bibliography{krr,procs}
\appendix
\section{Proofs}
\begin{lemma}\label{ht_model_correspondence_1}
	Let $\psi$ and $\varphi$ be formulas over \At ,  $\auxatom{\varphi} \not \in \At$ and $H \subseteq T \subseteq (\At \cup \{\auxatom{\varphi}\})$.
	Then, $\tuple{H,T} \models (\psi[\varphi/\auxatom{\varphi}] \wedge (\varphi \leftrightarrow \auxatom{\varphi}))$ if and only if $\tuple{H \cap \At,T \cap \At} \models \psi$ and $H\backslash \At = \{ x_\varphi \mid \tuple{H \cap \At,T \cap \At} \models \varphi \}$.
\end{lemma}

\begin{proof}
	Since~$x_\varphi \notin \At$, $\varphi$ and~$\psi$ are formulas over~$\At$, (by structural induction) we get that $\tuple{H,T} \models \varphi$ iff $\tuple{H\cap \At,T\cap \At} \models \varphi$ and $\tuple{H,T} \models \psi$ iff $\tuple{H\cap \At,T\cap \At} \models \psi$.
	
	$\implies$	Suppose $\tuple{H,T} \models (\psi[\varphi/\auxatom{\varphi}] \wedge (\varphi \leftrightarrow \auxatom{\varphi}))$.
	By the rule substitution of equivalents, we get $\tuple{H,T} \models (\psi \wedge (\varphi \leftrightarrow \auxatom{\varphi}))$ and it follows that $\tuple{H,T} \models \psi$ and $\tuple{H\cap \At,T\cap \At} \models \psi$.
	So, if  $\tuple{H,T} \models (\psi[\varphi/\auxatom{\varphi}] \wedge (\varphi \leftrightarrow \auxatom{\varphi}))$, then $\tuple{H\cap \At,T\cap \At} \models \psi$.
	
	Also, since $\tuple{H,T} \models (\varphi \leftrightarrow \auxatom{\varphi})$, we know that $\auxatom{\varphi} \in H$ iff  $\tuple{H,T} \models \varphi$.
	Consequently, $\auxatom{\varphi} \in H \backslash \At$ iff  $\tuple{H\cap \At,T\cap \At} \models \varphi$.
	It follows that if $\tuple{H,T} \models (\psi[\varphi/\auxatom{\varphi}] \wedge (\varphi \leftrightarrow \auxatom{\varphi}))$, then  $\tuple{H \cap \At,T \cap \At} \models \psi$ and $H\backslash \At = \{ x_\varphi \mid \tuple{H \cap \At,T \cap \At} \models \varphi \}$.
	
	$\impliedby$	Suppose $\tuple{H\cap \At,T\cap \At} \models \psi$ and $H\backslash \At = \{ x_\varphi \mid \tuple{H \cap \At,T \cap \At} \models \varphi \}$.
	Then $\tuple{H,T} \models \psi$, and $H\backslash \At = \{ x_\varphi \mid \tuple{H,T} \models \varphi \}$.
	It implies that $\tuple{H,T} \models (\psi \wedge (\varphi \leftrightarrow \auxatom{\varphi}))$ and by the rule substitution of equivalents, we get $\tuple{H,T} \models \psi[\varphi/x_\varphi] \wedge (\varphi \leftrightarrow x_\varphi)$.
\end{proof}

\begin{proof}[Proposition $\ref{prop:extension:definition}$]
	$\implies$	Let a set of atoms $T$ over \At\ be a stable model of $\psi$.
	Then $\tuple{T,T} \models \psi$.
	Consider a set of atoms $T' = T \cup \{ x_\varphi \mid \tuple{T,T} \models \varphi \}$.
	Then from Lemma $\ref{ht_model_correspondence_1}$, $\tuple{T',T'} \models \psi[\varphi/\auxatom{\varphi}] \wedge (\varphi \leftrightarrow \auxatom{\varphi})$.

	Suppose some interpretation $\tuple{H',T'}$ where $H' \subset T'$, is a model of $\psi[\varphi/\auxatom{\varphi}] \wedge (\varphi \leftrightarrow \auxatom{\varphi})$.
	Then, from Lemma $\ref{ht_model_correspondence_1}$, $((H' \cap \At) \cup \{\auxatom{\varphi} \mid \tuple{H' \cap \At, T} \models \varphi\}) \subset (T \cup\{\auxatom{\varphi} \mid \tuple{T, T} \models \varphi\})$ and $\tuple{H' \cap \At,T} \models \psi$.
	But since $T$ is a stable model of $\psi$, we get that $H' \cap \At = T$, this leads us to a contradiction.
	So $T'$ is a stable model $(\psi[\varphi/\auxatom{\varphi}] \wedge (\varphi \leftrightarrow \auxatom{\varphi}))$.
		
	$\impliedby$	Let a set of atoms $T'$ over $\At \cup\{\auxatom{\varphi} \}$ be a stable model of $(\psi[\varphi/\auxatom{\varphi}] \wedge (\varphi \leftrightarrow \auxatom{\varphi}))$. 
	Then $\tuple{T',T'} \models (\psi[\varphi/\auxatom{\varphi}] \wedge (\varphi \leftrightarrow \auxatom{\varphi}))$.
	Suppose $T = T' \cap \At$.
	Then from Lemma $\ref{ht_model_correspondence_1}$, $\tuple{T,T} \models \psi$ with  and $T'\backslash \At$ = $\{ x_\varphi \mid \tuple{T,T} \models \varphi \}$.

	Suppose some interpretation $\tuple{H,T}$ where $H \subset T$, is a model of $\psi$.
	Also, $H' = H \cup \{\auxatom{\varphi} \mid \tuple{H, T} \models \varphi\} \subset T \cup \{\auxatom{\varphi} \mid \tuple{H, T} \models \varphi\}$.
	By persistence, if $\tuple{H,T} \models \varphi$, then  $\tuple{T,T} \models \varphi$.
	So $H' \subset T'$
	and by Lemma $\ref{ht_model_correspondence_1}$,  $\tuple {H',T'} \models (\psi[\varphi/\auxatom{\varphi}] \wedge (\varphi \leftrightarrow \auxatom{\varphi}))$.
	But $T'$ is a stable model of $(\psi[\varphi/\auxatom{\varphi}] \wedge (\varphi \leftrightarrow \auxatom{\varphi}))$ and we get a contradiction.
	So $T$ is a stable model of $\psi$.

	It follows that $\psi$ and $(\psi[\varphi/\auxatom{\varphi}] \wedge (\varphi \leftrightarrow \auxatom{\varphi}))$ over $\At$ and hence $\psi \cong_{\At} (\psi[\varphi/\auxatom{\varphi}] \wedge (\varphi \leftrightarrow \auxatom{\varphi}))$.
	 
\end{proof}

\begin{proof}[Proposition \ref{prop:extension:definition:one:one}]
	It follows from the proof of Proposition $\ref{prop:extension:definition}$.
\end{proof}

\begin{lemma}\label{entail_proof}
	Let $\phi_1$ and $\phi_2$ be two formulas over any vocabulary.
	Then,
	\[
		(\neg \phi_1 \leftrightarrow \phi_2) \vdash (\neg \phi_2 \vee  \phi_2)
	\]
\end{lemma}
\begin{proof}
\[
	\begin{nd}
	\hypo {1} {\neg \phi_1 \leftrightarrow \phi_2}
	\have {2} {\neg \phi_1 \leftarrow \phi_2} \dae{1}
	\have {3} {\phi_2 \leftarrow \neg \phi_1} \dae{1}
	\have {4} {\neg \phi_1 \vee \neg \neg \phi_1} \wem{}
	\open
	\hypo {5} {\neg \phi_1}
	\have {6} {\phi_2}	\ie{3,5}
	\have {7} {\phi_2 \vee \neg \phi_2} \oi{6}
	\close
	\open
	\hypo {8} {\neg \neg \phi_1}
	\open
	\hypo {9} {\phi_2}
	\have {10} {\neg \phi_1} \ie{2,9} 
	\have {11} {\bot} \bi{8,10}
	\close
	\have {12} {\neg \phi_2} \ni{9-11}
	\have {13} {\neg \phi_2 \vee \phi_2} \oi{12}
	\close
	\have {14} {\neg \phi_2 \vee \phi_2} \oe{4}
	\end{nd}
\]
\end{proof}

\begin{lemma}\label{formula_se}
		For any pair of formulas $\phi_1$ and $\phi_2$, the following conditions hold:
		\begin{enumerate}
			\item
				$(\neg (\phi_1 \wedge \phi_2) \wedge \neg(\neg \phi_1 \wedge \neg \phi_2)) \cong \ (\neg \neg (\neg \phi_1 \leftrightarrow \phi_2))$
			\item
				$(\neg \neg (\neg \phi_1 \leftrightarrow \phi_2) \wedge (\neg \phi_2 \vee  \phi_2)) \cong \  ((\neg \phi_1 \leftrightarrow \phi_2) \wedge (\neg \phi_2 \vee  \phi_2))$
		\end{enumerate}
\end{lemma}

\begin{proof}
	For any two formulas $\phi_1$ and $\phi_2$ over \At,

			First, we prove that
			\begin{equation}\label{or_implication}
				(\phi_1 \vee \neg \phi_2) \vdash (\phi_1 \leftarrow \phi_2)
			\end{equation}
			\[
	\begin{nd}
	\hypo {1} {\phi_1\vee \neg \phi_2}
	\open
	\hypo {2} {\phi_2}
	\open
	\hypo {3} {\phi_1}
	\have {4} {\phi_1} \r{3}
	\close
	\open
	\hypo {5} {\neg \phi_2}
	\have {6} {\bot} \ne{2,5}
	\have {7} {\phi_1} \be{7}
	\close
	\have {8} {\phi_1} \oe{1,3-4,5-7}
	\close
	\have {9} {\phi_1 \leftarrow \phi_2} \ii{2-8}
	\end{nd}
\]

			Then, we can prove that,
			\begin{equation}\label{formula_se_1_f}
				(\neg(\phi_1\wedge \phi_2) \wedge \neg(\neg \phi_1\wedge \neg \phi_2)) \vdash  \neg \neg(\neg \phi_1\leftrightarrow \phi_2)
			\end{equation}
			\[
	\begin{nd}
	\hypo {1} {\neg(\phi_1\wedge \phi_2) \wedge \neg(\neg \phi_1\wedge \neg \phi_2)}
	\have {2} {\neg(\phi_1\wedge \phi_2)} \ae{1}
	\have {3} {\neg \phi_1\vee \neg \phi_2}	\dem{2}
	\have {4} {\neg \phi_1\leftarrow \phi_2} \pr1{3}
	\have {5} {\neg \neg(\neg \phi_1\leftarrow \phi_2) } \dni{4}
	\have {6} {\neg(\neg \phi_1\wedge \neg \phi_2)} \ae{1}
	\have {7} {\neg \neg (\phi_2\vee \phi_1)} \dem{6}
	\have {8} {\neg \neg (\phi_2\vee \neg \neg \phi_1)} \dni{7}
	\have {9} {\neg \neg (\phi_2\leftarrow \neg \phi_1)}  \pr1{8}
	\have {10} {\neg \neg(\neg \phi_1\leftarrow \phi_2) \wedge \neg \neg (\phi_2\leftarrow \neg \phi_1)} \ai{5,9}
	\have {11} {\neg (\neg(\neg \phi_1\leftarrow \phi_2) \vee \neg  (\phi_2\leftarrow \neg \phi_1))} \dem{10}
	\have {12} {\neg \neg((\neg \phi_1\leftarrow \phi_2) \wedge (\phi_2\leftarrow \neg \phi_1))} \dem{11}
	\have {13} {\neg \neg(\neg \phi_1\leftrightarrow \phi_2)} \dai{12}
	\end{nd}
\]

			We prove that,
			\begin{equation}\label{formula_se_1_b}
				\neg \neg(\neg \phi_1\leftrightarrow \phi_2)\vdash (\neg(\phi_1\wedge \phi_2) \wedge \neg(\neg \phi_1\wedge \neg \phi_2))
			\end{equation}
			\[
	\begin{nd}
	\hypo {1} {\neg \neg(\neg \phi_1\leftrightarrow \phi_2)}
	\open
	\hypo {2} {\phi_1 \wedge \phi_2}
	\open
	\hypo {3} {\neg \phi_1 \leftrightarrow \phi_2}
	\have {4} {\neg \phi_1 \leftarrow \phi_2} \dae{3}
	\have {5} {\phi_2} \ae{2}
	\have {6} {\neg \phi_1} \ie{3,4}
	\have {7} {\phi_1} \ae{2}
	\have {8} {\bot} \bi{6,7}
	\close
	\have {9} {\neg(\neg \phi_1 \leftrightarrow \phi_2)} \be{3-8}
	\have {10} {\bot} \bi{1,9}
	\close
	\have {11} {\neg(\phi_1 \wedge \phi_2)} \be{2-10}
	\open
	\hypo {12} {\neg \phi_1 \wedge \neg \phi_2}
	\open
	\hypo {13} {\neg \phi_1 \leftrightarrow \phi_2}
	\have {14} {\neg \phi_1 \leftarrow \phi_2} \dae{13}
	\have {15} {\neg \phi_1} \ae{12}
	\have {16} {\phi_2} \ie{13,14}
	\have {17} {\neg \phi_2} \ae{12}
	\have {18} {\bot} \bi{16,17}
	\close
	\have {19} {\neg(\neg \phi_1 \leftrightarrow \phi_2)} \be{13-18}
	\have {20} {\bot} \bi{11,19}
	\close
	\have {21} {\neg(\neg \phi_1 \wedge \neg \phi_2)} \be{12-20}
	\have {22} {(\neg(\phi_1\wedge \phi_2) \wedge \neg(\neg \phi_1\wedge \neg \phi_2))} \ai{11,21}
	\close
	\end{nd}
\]

			So from (\ref{formula_se_1_f}) and (\ref{formula_se_1_b}),
			\begin{equation}\label{formula_se_1}
				(\neg (\phi_1 \wedge \phi_2) \wedge \neg(\neg \phi_1 \wedge \neg \phi_2)) \cong (\neg \neg (\neg \phi_1 \leftrightarrow \phi_2))
			\end{equation}
			We prove that,
			\begin{equation}\label{formula_se_2_f}
			(\neg(\phi_1\wedge \phi_2) \wedge \neg(\neg \phi_1\wedge \neg \phi_2) \wedge (\neg  \phi_2\vee \phi_2)) \vdash (\neg \phi_1\leftrightarrow \phi_2)
			\end{equation}
			\[
	\begin{nd}
	\hypo {1} {\neg(\phi_1\wedge \phi_2) \wedge \neg(\neg \phi_1\wedge \neg \phi_2)\wedge (\neg  \phi_2\vee \phi_2)}
	\have {2} {\neg(\phi_1\wedge \phi_2)} \ae{1}
	\have {3} {\neg\phi_1\vee \neg\phi_2)} \dem{2}
	\have {4} {\neg\phi_1\leftarrow \phi_2)} \pr1{3}
	\have {5} {\neg(\neg \phi_1\wedge \neg \phi_2)} \ae{1}
	\have {6} {\neg \neg \phi_1\vee \neg \neg \phi_2)} \dem{5}
	\have {7} {\neg \neg \phi_2\leftarrow \neg \phi_1)} \pr1{6}
	\have {8} {\neg  \phi_2\vee \phi_2} \ae{1}
	\open
	\hypo {9} {\neg \phi_2}
	\have {10} {\neg \neg \neg  \phi_2} \tni{9}
	\have {11} {\neg \neg \neg  \phi_2 \vee \phi_2} \oi{10}
	\close
	\open
	\hypo {12} {\phi_2}
	\have {13} {\phi_2} \r{12}
	\have {14} {\phi_2 \vee \neg \neg \neg  \phi_2} \oi{13}
	\close
	\have {15} {\phi_2 \vee \neg \neg \neg  \phi_2} \oe{8,9-11,12-14}
	\have {16} {\phi_2 \leftarrow \neg \neg \phi_2} \pr1{15}
	\have {17} {\phi_2 \leftarrow \neg \phi_1} \ie{7,16}
	\have {18} {\neg \phi_1 \leftrightarrow \phi_2 } \dai{4,17}
	\end{nd}
\]

			Then from (\ref{formula_se_1}) and (\ref{formula_se_2_f}),
			\begin{equation}\label{formula_se_2_f1}
			(\neg \neg(\neg \phi_1\leftrightarrow \phi_2) \wedge (\neg \phi_2\vee  \phi_2)) \vdash (\neg \phi_1\leftrightarrow \phi_2)
			\end{equation}
			We know that,
			\begin{equation}\label{formula_se_2_b}
			(\neg \phi_1\leftrightarrow \phi_2)	 \vdash (\neg \neg(\neg \phi_1\leftrightarrow \phi_2))
			\end{equation}
			So from (\ref{formula_se_2_f1}) and (\ref{formula_se_2_b}),
			\begin{equation}\label{formula_se_2}
			(\neg \neg (\neg \phi_1 \leftrightarrow \phi_2) \wedge (\neg \phi_2 \vee  \phi_2)) \cong ((\neg \phi_1 \leftrightarrow \phi_2) \wedge (\neg \phi_2 \vee  \phi_2))
			\end{equation}
\end{proof}

\begin{proof}[Theorem $\ref{thm:extension:definition:negated}$]
	Let $\negocc{\psi} = \{\varphi_1,...,\varphi_n\}$.
	For $1\leq i \leq n$, we define a sequence of formulas as:
	\begin{align}
		\psi_0 =& \psi\\
		\psi_i =& \psi_{i-1} \wedge (\varphi_i\leftrightarrow\auxatom{\varphi_i})
	\end{align}
	Then $\psi_n = \psi \wedge	 \bigwedge_{\varphi\in\negocc{\psi}}(\varphi\leftrightarrow\auxatom{\varphi})$.
	From Proposition $\ref{prop:extension:definition}$, we know that $\psi \cong_{\At} (\psi[\varphi_1/\auxatom{\varphi_1}] \wedge (\varphi_1\leftrightarrow\auxatom{\varphi_1}))$ and on substitution of the equivalents, we get  $\psi \cong_{\At} (\psi \wedge (\varphi_1\leftrightarrow\auxatom{\varphi_1}))$ and it follows that	 $\psi_0 \cong_{\At} \psi_1$.
	So, for $1\leq i \leq n$, $\psi_{i-1} \cong_{\At} \psi_{i}$ and $\psi_0 \cong_{\At} \psi_n$, that is,
	\[\psi
	\cong_{\At}
	(\psi \wedge	 \bigwedge_{\varphi\in\negocc{\psi}}(\varphi\leftrightarrow\auxatom{\varphi}))
	\]
	By substitution of equivalents,
	\[\psi
	\cong_{\At}
	\psi_n
	\cong
	(\psi\big[\varphi/\auxatom{\varphi}\mid\varphi\in\negocc{\psi}\big] \wedge	 \bigwedge_{\varphi\in\negocc{\psi}}(\varphi\leftrightarrow\auxatom{\varphi}))
	\]

	For $\varphi\in\negocc{\psi}$, from Lemma $\ref{entail_proof}$, we can see that, $	( \varphi \leftrightarrow \auxatom{\varphi}) \vdash (\neg \auxatom{\varphi} \vee  \auxatom{\varphi})$.
	So,
	\[\psi
	\cong_{\At}
	\psi_n
	\cong (\psi\big[\varphi/\auxatom{\varphi}\mid\varphi\in\negocc{\psi}\big] \wedge	 \bigwedge_{\varphi\in\negocc{\psi}}(\varphi\leftrightarrow\auxatom{\varphi}) \wedge \bigwedge_{\varphi\in\negocc{\psi}}(\neg\auxatom{\varphi}\vee\auxatom{\varphi}))\]
	From Lemma $\ref{formula_se}$ part 2,
	\[
	((\varphi \leftrightarrow \auxatom{\varphi}) \wedge (\neg \auxatom{\varphi} \vee \auxatom{\varphi})) \cong  (\neg \neg ( \varphi \leftrightarrow \auxatom{\varphi}) \wedge (\neg \auxatom{\varphi} \vee  \auxatom{\varphi}))
	\]
	It follows that,
	\[\psi
	\cong_{\At}
	\psi_n
	\cong
	(\psi\big[\varphi/\auxatom{\varphi}\mid\varphi\in\negocc{\psi}\big] \wedge	 \bigwedge_{\varphi\in\negocc{\psi}}\neg \neg (\varphi\leftrightarrow\auxatom{\varphi}) \wedge \bigwedge_{\varphi\in\negocc{\psi}}(\neg\auxatom{\varphi}\vee\auxatom{\varphi}))= \translation{\psi} \]
	Hence,
	\(\psi 	\cong_{\At} \translation{\psi}\).
\end{proof}

\begin{proof}[Theorem \ref{thm:extension:definition:negated:one:one}]
	Let $\negocc{\psi} = \{\varphi_1,...,\varphi_n\}$.
	Similar to proof of Theorem $\ref{thm:extension:definition:negated}$, for $1\leq i \leq n$, we define a sequence of formulas as:
	\begin{align}
	\psi_0 =& \psi\\
	\psi_i =& \psi_{i-1} \wedge (\varphi_i\leftrightarrow\auxatom{\varphi_i})
	\end{align}
	Then $\psi_n = \psi \wedge	 \bigwedge_{\varphi\in\negocc{\psi}}(\varphi\leftrightarrow\auxatom{\varphi})$.

	For $1\leq i \leq n$, we define a sequence of set of atoms as:
	\begin{align}
	T_0 =& T\\
	T_i =& T_{i-1} \cup \{\auxatom{\varphi_i} \mid \tuple{T ,T} \models \varphi_i \}
	\end{align}
	Then, $T_n = T \cup \{\auxatom{\varphi} \mid \varphi \in \negocc{\psi}, \tuple{T ,T} \models \varphi \}$.
	By the substitution of equivalents, we can see that, $\psi_1 \cong (\psi\big[\varphi_1/\auxatom{\varphi_1}\big] \wedge (\varphi_1\leftrightarrow\auxatom{\varphi_1}))$.
	Additionally, from Proposition $\ref{prop:extension:definition:one:one}$ we get that $T$ is a stable model of $\psi$ if and only if  $T \cup \{\auxatom{\varphi_1} \mid \tuple{T ,T} \models \varphi_1 \}$ is a stable model of $(\psi\big[\varphi_1/\auxatom{\varphi_1}\big] \wedge (\varphi_1\leftrightarrow\auxatom{\varphi_1}))$.
	It follows that $T$ is a stable model of $\psi$ if and only if $T_1$ is a stable model of $\psi_1$.
	Similarly, for $1\leq i \leq n$, $T_{i-1}$ is a stable model of $\psi_{i-1}$ if and only if $T_i$ is a stable model of $\psi_i$.
	It follows that $T$ is a stable model of $\psi$ if and only if $T_n$ is a stable model of $\psi_n$.
	From the proof of Theorem $\ref{thm:extension:definition:negated}$, we know that \(\psi_n \cong \translation{\psi}\). It follows that $T$ is a stable model of $\psi$ if and only if $T_n$ is a stable model of $\translation{\psi}$ and hence the Theorem's statements follows.
\end{proof}

\begin{proof}[Corollary \ref{cor:extension:definition:program}]
	From part 1 of Lemma $\ref{formula_se}$, we can see that $ (\neg \neg (\neg a \leftrightarrow \auxatm{a}))$ can be expressed in form of integrity constraints as
	\begin{align*}
		\leftarrow&       a\wedge     \auxatm{a} \\
		\leftarrow&  \neg a\wedge \neg\auxatm{a}
	\end{align*}

	Also, the disjunction $\auxatm{a} \vee \neg \auxatm{a}$ is a choice rule of form $\{\auxatm{a}\} \leftarrow$.
	So from Theorem $\ref{thm:extension:definition:negated}$, $r \cong_{\At} \translation{r}$ and $P \cong_{\At} \translation{P}$.
\end{proof}

\begin{proof}[Corollary \ref{cor:extension:definition:program:one:one}]
	From the proof of  Corollary \ref{cor:extension:definition:program}, we can see that for a rule $r$, $\translation{r}$ using Definition \ref{def:translation} can be expressed alternatively as shown in Definition \ref{def:rule:translation}.
	So, the Corollary follows from program translation in Definition \ref{def:program:translation} and from Theorem \ref{thm:extension:definition:negated:one:one}.
\end{proof}

\begin{proof}[Proposition \ref{prop:austere:stable:models}]
	$\implies$ Let a set of atoms $X$ be a stable model of $(F,D,C,I)$.
	Suppose $C' \subseteq C$ be the set of choice rules of form 
	\begin{equation} \label{choice_rule}
		\{a\} \leftarrow
	\end{equation}
	where $a \in X$. 
	We know $\tuple{X,X} \models (F \cup C' \cup D)$.
	Also, a rule $r \in (F\cup F_{C'} \cup D)$ can be expressed as
	\begin{equation} \label{definite_rule}
	a \leftarrow B 
	\end{equation}
	where $a$ is an atom and $B$ is a set of atoms.
	It follows that $a \in X$ if $B \subseteq  X$ and $T_{(F\cup F_{C'} \cup D)}(X) \subseteq X$.

	Suppose $X'$ is a pre-fixpoint of $T_{(F\cup F_{C'} \cup D)}$ where $X' \subset X$.
	So $T_{(F\cup F_{C'} \cup D)}(X') \subseteq X'$ and then for a rule $r \in (F\cup F_{C'} \cup D)$ expressed as $(\ref{definite_rule})$, $a \in X'$ if $B \subseteq X'$ if $B \subseteq X$.
	It follows that $\tuple{X',X} \models (F\cup F_{C'} \cup D)$.  
	Since $\tuple{X',X} \models F_{C'}$, then for a rule $r  \in F_{C'}$ of form 
	\begin{equation} \label{fact}
		\{a\} \leftarrow
	\end{equation}
	 we see that $a \in X'$, $\tuple{X',X} \models a$, $\tuple{X',X} \models (a; \neg a)$ and hence  $\tuple{X',X} \models C'$.
	For $r \in C\backslash C'$ of form (\ref{choice_rule}), we know that $a \not \in X$, $\tuple{X,X} \not\models a$, $\tuple{X',X} \models \neg a$, $\tuple{X',X} \models (a; \neg a)$ and hence  $\tuple{X',X} \models C\backslash C'$.
	So $\tuple{X',X} \models (F \cup C \cup D \cup I)$ but since $X$ is the stable model of $(F \cup C \cup D \cup I)$, we get a contradiction.
	It follows that $X$ is the least fixpoint of $T_{(F \cup F_{C'} \cup D)}$ where $C' \subseteq C$. 
	So $X$ is a candidate stable model of $(F,C,D,I)$ and $\tuple{X,X} \models I$.   
	
	$\impliedby$
	Suppose a set of atoms $X$ be a candidate stable model of $(F,D,C,I)$ such that $\tuple{X,X} \models I$. Then, $X$ is the least fixpoint of $T_{(F\cup F_{C'} \cup D)}$ for some $C' \subseteq C$ and $T_{(F\cup F_{C'} \cup D)}(X) \subseteq X$.
	Any rule $r \in (F \cup F_{C'} \cup D)$ is of the form ($\ref{definite_rule}$) and it follows that $a \in X$ if $B \subseteq X$.
	So $\tuple{X,X} \models r$ and we get that $\tuple{X,X} \models (F \cup F_{C'} \cup D)$.
	For a rule $r \in C$ of form ($\ref{choice_rule}$), $a \in X$ or $a \not \in X$. 
	So  $\tuple{X,X} \models (a;\neg a)$ and $\tuple{X,X} \models r$.
	It follows that $\tuple{X,X} \models (F \cup C \cup D \cup I)$.

	Suppose some $\tuple{X',X} \models (F \cup C \cup D \cup I)$, where $X' \subset X$.
	Then for $r \in (F \cup F_{C'} \cup D)$, $a \in X'$ if $\tuple{X',X} \models B$ if $B \subseteq X'$.
	Hence $T_{(F\cup F_{C'} \cup D)} (X') \subseteq X'$.
	Since $X$ is the least pre-fixpoint of $T_{(F\cup F_{C'} \cup D)}$, we arrive at a contradiction and so $X$ is the stable model of $(F,D,C,I)$.
\end{proof}

\end{document}